\documentclass[11pt,a4paper]{article}
\usepackage[hyperref]{emnlp-ijcnlp-2019}
\usepackage{times}
\usepackage{latexsym}
\usepackage{multirow}
\usepackage{longtable}
\usepackage{tabularx}

\usepackage{url}

\aclfinalcopy 

\input{Definitions.tex}

\title{Robustness to Capitalization Errors in Named Entity Recognition}

\author{Sravan Bodapati \\
  Amazon.com \\
  \texttt{sravanb@amazon.com} \\\And
  Hyokun Yun \\
  Amazon.com \\
  \texttt{yunhyoku@amazon.com} \\\And
  Yaser Al-Onaizan \\
  Amazon.com \\
  \texttt{onaizan@amazon.com}}
\date{}

\begin{document}
\maketitle
\begin{abstract}
  Robustness to capitalization errors is a highly desirable characteristic
  of named entity recognizers, yet we find standard models for the task are
  surprisingly brittle to such noise. 
  Existing methods to improve robustness to the noise
  completely discard given orthographic information,
  which significantly degrades their performance on well-formed text.
  We propose a simple alternative approach based on data augmentation,
  which allows the model to \emph{learn} to utilize or ignore orthographic information
  depending on its usefulness in the context. It achieves competitive robustness 
  to capitalization errors 
  while making negligible compromise to its performance on well-formed text
  and significantly improving generalization power on noisy user-generated text.
  Our experiments clearly and consistently 
  validate our claim across different types of machine learning models, languages, 
  and dataset sizes.
\end{abstract}

\section{Introduction}

In the last two decades, substantial progress has been made on the
task of named entity recognition (NER), as it has enjoyed the 
development of probabilistic modeling
\citep{lafferty2001conditional, finkel2005incorporating},
methodology \citep{ratinov2009design},
deep learning \citep{collobert2011natural, huang2015bidirectional, lample2016neural}
as well as semi-supervised learning \citep{peters2017semi, peters2018deep}.
Evaluation of these developments, however, has been mostly focused on their
impact on global average metrics, most notably the micro-averaged F1 score \citep{chinchor1992muc}.

For practical applications of NER, however,
there can be other considerations for model evaluation. 
While standard training
data for the task consists mainly of well-formed text \citep{tksintro2002conll, pradhan2009ontonotes}, 
models trained on such data are often applied on a broad
range of domains and genres by users who are not necessarily NLP experts, thanks to the proliferation of
toolkits \citep{manning2014stanford} and general-purpose machine learning services.
Therefore, there is an increasing demand for the strong robustness of models to 
unexpected noise.

In this paper, we tackle 
one of the most common types of noise in applications of NER: unreliable capitalization.
Noisiness in capitalization is a typical characteristic of user-generated text 
\citep{ritter2011named, baldwin2015shared}, but it is not uncommon even 
in formal text. Headings, legal documents, or emphasized sentences are often
capitalized.
All-lowercased text, on the other hand, 
can be produced in large scale from upstream machine learning models
such as speech recognizers and machine translators \citep{kubala1998named},
or processing steps in the data pipeline which are not fully under the control of the 
practitioner.  Although a text without correct capitalization is perfectly legible for human readers
\citep{cattell1886time, rayner1975perceptual}
with only a minor impact on the reading speed \citep{tinker1928influence, arditi2007letter}, 
we show that typical NER models are surprisingly brittle to 
all-uppercasing or all-lowercasing of text.
The lack of robustness these models show to
such common types of noise makes them unreliable, especially when 
characteristics of target text are not known a priori.

\begin{table*}[ht]
  \centering
  {\small
  \begin{tabular}{cc|cccccc}
    \hline \hline
    & Annotation & \texttt{O} & \texttt{O} & \texttt{O} & \texttt{B-ORG} & \texttt{I-ORG} & \texttt{E-ORG} \\
    \hline
    (a) & Original Sentence & I & live & in & New & York & City \\
    (b) & Lower-cased Sentence & i & live & in & new & york & city \\
    (c) & Upper-cased Sentence & I & LIVE & IN & NEW & YORK & CITY \\
    \hline \hline
  \end{tabular}
  }
  \caption{Example of Data Augmentation}
  \label{tab:example}
\end{table*}

There are two standard treatments on the problem in the literature.
The first is to train a case-agnostic model \citep{kubala1998named, robinson1999overview}, 
and the second is to explicitly
correct the capitalization \citep{srihari2003case, lita2003truecasing, ritter2011named}.
One of the main contributions of this paper is to empirically evaluate
the effectiveness of these techniques across models, languages, and dataset sizes.
However, both approaches have clear conceptual limitations.
Case-agnostic models discard orthographic information (how the
given text was capitalized),
which is considered to be highly useful \citep{robinson1999overview}; 
our experimental results also support this. The second approach of 
correcting the capitalization of the text, on the other hand, requires
an access to a high-quality truecasing model, and errors from
the truecasing model would cascade to final named entity predictions.

We argue that an ideal approach should take a full advantage of orthographic information
when it is correctly present, but rather than assuming the information to be
always perfect, the model should be able to \emph{learn} to ignore the orthographic information when 
it is unreliable.
To this end, we propose a novel approach based on 
data augmentation \citep{simard2003best}. In computer vision, data augmentation
is a highly successful standard technique \citep{krizhevsky2012imagenet},
and it has found adoptions in natural language processing tasks such as text classification
\citep{zhang2015text}, question-answering \citep{yu2018qanet} and low-resource learning \citep{sahin2018data}. 
Consistently across
a wide range of models (linear models, deep learning models to deep contextualized models), 
languages (English, German, Dutch, and Spanish), and dataset sizes (CoNLL 2003 and OntoNotes 5.0),
the proposed method shows strong robustness while making little compromise to the performance
on well-formed text.

\section{Formulation}

Let $\xb=(x_1, x_2, \ldots, x_n)$ be a sequence of words in a sentence.
We follow the standard approach of formulating NER as a sequence tagging task
\citep{rabiner1989tutorial, lafferty2001conditional, collins2002discriminative}.
That is, we predict a sequence of tags $\yb=(y_1, y_2, \ldots, y_n)$ 
where each $y_i$ identifies the type of the entity the word $x_i$ belongs to,
as well as the position of it in the surface form according to 
IOBES scheme \citep{uchimoto2000named}.
See Table~\ref{tab:example} (a) for an example annotated sentence.
We train probabilistic models under the maximum likelihood principle,
which
produce a probability score $\PP\sbr{\yb \mid \xb}$ for any possible
output sequence $\yb$.

All-uppercasing and all-lowercasing are common types of capitalization
errors.
Let $\text{upper}(x_i)$ and $\text{lower}(x_i)$ be functions that
lower-cases and upper-cases the word $x_i$, respectively. 
Robustness of a probabilistic model to these types of noise can be understood
as the quality of scoring function 
$\PP[\yb \mid \text{upper}(x_1), \ldots, \text{upper}(x_n)]$
and $\PP[\yb \mid \text{lower}(x_1), \ldots, \text{lower}(x_n)]$
in predicting the correct annotation $\yb$, which can still be quantified
with standard evaluation metrics such as the micro-F1 score. 

\section{Prior Work}

There are two common strategies to improve robustness to capitalization errors.
The first is to completely ignore orthograhpic information by using case-agnostic models
\citep{kubala1998named, robinson1999overview}. For linear models, this can be achieved by
restricting the choice of features to case-agnostic ones. On the other hand, deep learning models 
without hand-curated features \citep{lample2016neural, chiu2016named}
can be easily made case-agnostic by lower-casing every input to the model.
The second strategy is to explictly correct the capitalization by using another model trained for
this purpose,  which is called ``truecasing''\citep{srihari2003case, lita2003truecasing}.
Both methods, however, have the common limitation that they discard orthographic information 
in the target text, which can be correct; this leads to degradation of performance on well-formed text.

\begin{table*}[ht]
    \centering
    {\small
    \begin{tabular}{cc|ccc|ccc|ccc}
    \hline \hline
    \multirow{2}{*}{Model} & \multirow{2}{*}{Method} & \multicolumn{3}{c|}{CoNLL-2003 English} & \multicolumn{3}{c|}{OntoNotes 5.0 English} & \multicolumn{3}{c}{Transfer to Twitter}\\
    \cline{3-11}
    & & Original & Lower & Upper & Original & Lower & Upper & Original & Lower & Upper \\ \hline
    \multirow{4}{*}{Linear} & Baseline & 89.2 & 57.8 & 75.2 & 81.7 & 37.4 & 15.1 & 24.4 & 6.9 & 20.2 \\
    & Caseless & 83.7 & 83.7 & 83.7 & 75.5 & 75.5 & 75.5 & 20.3 & 20.3 & 20.3 \\
    & Truecasing & 83.8 & 83.8 & 83.8 & 76.6 & 76.6 & 76.6 & 24.0 & 24.0 & 24.0 \\
    & DA & 88.2 & 85.6 & 86.1 & - & - & - & 28.2 & 26.4 & 27.0 \\ \hline 
    \multirow{4}{*}{BiLSTM} & Baseline & 90.8 & 0.4 & 52.3 & 87.6 & 38.9 & 15.5 & 18.1 & 0.1 & 7.9\\
    & Caseless & 85.7 & 85.7 & 85.7 & 83.2 & 83.2 & 83.2 & 20.3 & 20.3 & 20.3 \\
    & Truecasing & 84.6 & 84.6 & 84.6 & 81.7 & 81.7 & 81.7 & 18.7 & 18.7 & 18.7\\
    & DA & 90.4 & 85.3 & 83.8 & 87.5 & 83.2 & 82.6 & 21.2 & 17.7 & 18.4 \\ \hline
    \multirow{4}{*}{ELMo} & Baseline & 92.0 & 34.8 & 71.6 & 88.7 & 66.6 & 48.9 & 31.6 & 1.5 & 19.6 \\
    & Caseless & 89.1 & 89.1 & 89.1 & 85.3 & 85.3 & 85.3 & 31.8 & 31.8 & 31.8 \\
    & Truecasing & 86.2 & 86.2 & 86.2 & 83.2 & 83.2 & 83.2 & 28.8 & 28.8 & 28.8\\
    & DA & 91.3 & 88.7 & 87.9 & 88.3 & 85.8 & 83.6 & 34.6 & 31.7 & 30.2 \\
    \hline \hline
    \end{tabular}
    }
\caption{F1 scores on original, lower-cased, and upper-cased test sets of English Datasets.
    Stanford Core NLP could not be trained on the augmented dataset even with 512GB of RAM.}
\label{tab:exp-eng}
\end{table*}

\section{Data Augmentation}
\label{sec:da}

Data augmentation refers to a technique of increasing the size of training data
by adding label-preserving transformations of them \citep{simard2003best}. For example, in
image classification, an object inside of an image does not change if the image is rotated,
translated, or slightly skewed; most people would still recognize the same object they would
find in the original image. By training a model on 
transformed versions of training images, the model becomes 
invariant to the transformations used \citep{krizhevsky2012imagenet}.

In order to improve the robustness of NER models to capitalization
errors, we appeal to the same idea. When a sentence is all-lowercased or all-uppercased 
as in Table~\ref{tab:example} (b) and (c), each word would still correspond to the same entity.
This implies such transformations
are also label-preserving ones: for a sentence $\xb$ and its ground-truth
annotation $\yb$, $\yb$ would still be a correct annotation for the all-uppercased sentence $\rbr{\text{upper}(x_1), \ldots, \text{upper}(x_n)}$
as well as the all-lowercased version 
$\rbr{\text{lower}(x_1), \ldots, \text{lower}(x_n)}$.
Indeed, all three sentences (a), (b) and (c) in Table~\ref{tab:example}
would share the same annotation.

\section{Experiments}


\begin{table*}
    \centering
    {\small
    \begin{tabular}{cc|ccc|ccc|ccc}
    \hline \hline
    \multirow{2}{*}{Model} & \multirow{2}{*}{Method} & \multicolumn{3}{c|}{CoNLL-2002 Spanish} & \multicolumn{3}{c|}{CoNLL-2002 Dutch} & 
    \multicolumn{3}{c}{CoNLL-2003 German}\\
    \cline{3-11}
    & & Original & Lower & Upper & Original & Lower & Upper & Original & Lower & Upper \\ \hline
    \multirow{3}{*}{Linear} & Baseline & 80.7 & 1.1 & 22.1 & 79.1 & 9.8 & 9.7 & 68.4 & 11.8 & 11.3 \\
    & Caseless & 69.9 & 69.9 & 69.9 & 63.9 & 63.9 & 63.9 & 53.3 & 53.3 & 53.3 \\
    & DA & 77.3 & 70.9 & 73.2 & 74.4 & 68.5 & 68.5 & 61.8 & 57.8 & 62.8 \\  \hline
    \multirow{3}{*}{BiLSTM} & Baseline & 85.4 & 1.0 & 26.8 & 87.3 & 2.0 & 15.8 & 79.5 & 6.5 & 9.8 \\
    & Caseless & 77.8 & 77.8 & 77.8 & 77.7 & 77.7 & 77.7 & 69.8 & 69.8 & 69.8 \\
    & DA & 85.3 & 78.4 & 76.5 & 84.8 & 75.0 & 75.9 & 76.8 & 69.7 & 69.7 \\ 
    \hline \hline
    \end{tabular}
    }
\caption{F1 scores on original, lower-cased, and upper-cased test sets of Non-English Datasets}
\label{tab:exp-noneng}
\end{table*}

We consider following three models,
each of which is state-of-the-art in their respective group:
\textbf{Linear}: Linear CRF model \citep{finkel2005incorporating} from
        Stanford Core NLP \citep{manning2014stanford}, which is representative of 
        feature engineering approaches.
        \textbf{BiLSTM}: Deep learning model from \citet{lample2016neural}
        which uses bidirectional LSTM
        for both character-level encoder and word-level encoder
        with CRF loss.  This is the state-of-the-art supervised 
        deep learning approach \citep{reimers2017reporting}.
        \textbf{ELMo}: 
        Bidirectional LSTM-CRF model which uses contextualized features from 
        deep bidirectional LSTM
        language model \citep{peters2018deep}. 
        For all models, we used hyperparameters from original papers.

We compare four strategies:
\textbf{Baseline}: Models are trained on unmodified training data.
    \textbf{Caseless}: We lower-case input data
        both at the training time and at the test time.
    \textbf{Truecasing}: Models are still trained on unmodified training data,
        but every input to test data is ``truecased'' \citep{lita2003truecasing} 
        using CRF truecasing model from Stanford Core NLP \citep{manning2014stanford}, which ignores given orthographic information in the text.  Due to the lack of access to truecasing models
        in other languages, this strategy was used only on English.
    \textbf{DA (Data Augmentation)}: We augment the original training set with
        upper-cased and lower-cased versions of it, 
        as discussed in Section~\ref{sec:da}.

We evaluate these models and methods on three versions of the test set for each dataset:
     \textbf{Original}: Original test data.
     \textbf{Upper}: All words are upper-cased.
     \textbf{Lower}: All words are lower-cased.
     Note that both Caseless and Truecasing method perform equally on all
     three versions because they ignore any original orthographic
    information in the \emph{test} dataset. We focus on micro-averaged F1 scores.


We use CoNLL-2002 Spanish and Dutch \citep{tksintro2002conll} and CoNLL-2003 English and German
\citep{sang2003introduction} to cover four languages, all of which orthographic information is useful in 
idenfitying named entities, and upper or lower-casing of text is straightforward.
We additionally evaluate on OntoNotes 5.0 English \citep{pradhan2009ontonotes}, which is
about five times larger than CoNLL datasets and contains more diverse genres.
F1 scores are shown in Table~\ref{tab:exp-eng} and \ref{tab:exp-noneng}.

\textbf{Question 1: How robust are NER models to
    capitalization errors?}
Models trained with the standard Baseline strategy suffer from significant
loss of performance when the test sentence is upper/lower-cased (compare `Original' column with `Lower' and `Upper').  For example, F1 score of BiLSTM on
lower-cased CoNLL-2003 English is abysmal 
0.4\%, completely losing any predictive power.
Linear and ELMo are more robust than BiLSTM thanks to smaller capacity
and semi-supervision respectively, but the degradation 
is still strong, ranging 20pp to 60pp loss in F1.

\textbf{Question 2: How effective Caseless, Truecasing, and Data Augmentation approaches
    are in improving robustness of models?}
All methods show similar levels of performance on lowercased or uppercased text.
Since Caseless and Data Augmentation strategy do not require additional
language-specifc resource as truecasing does, they seem to be superior to the
truecasing approach, at least on CoNLL-2003 English and OntoNotes 5.0 datasets with the
particular truecasing model used.
Across all datasets, the performance of Linear model on lower-cased or upper-cased
test set is consistently enhanced with data augmentation, compared with caseless models.


\textbf{Question 3: How much performance on well-formed text is sacrificed
    due to robustness?}
Caseless and Truecasing methods are perfectly robust to capitalization errors,
but only at the cost of significant degradation on well-formed text:
caseless and truecasing strategy lose 5.1pp
and 6.2pp respectively on the original test set of CoNLL-2003 English compared to
Baseline strategy, and on non-English datasets the drop is even bigger.
On the other hand, data augmentation preserves most of the performance on
the original test set: with BiLSTM, its F1 score drops by only 0.4pp and 0.1pp
respectively on CoNLL-2003 and OntoNotes 5.0 English. On non-English datasets,
the drop is bigger (0.1pp on Spanish but 2.5pp on Dutch and 2.7pp on German)
but still data augmentation performs about 7pp higher than Caseless
on original well-formed text across languages.

    

\textbf{Question 4: How do models trained on well-formed text generalize to noisy user-generated text?} 
The robustness of models is especially important when the characteristics of target text
are not known at the training time and can deviate significantly from those of training data.
To this end, we trained models on
CoNLL 2003-English, and evaluated them on annotations of Twitter data from \citet{fromreide2014crowdsourcing},
which exhibits natural errors of capitalization common in user-generated text.
`Transfer to Twitter' column of Table~\ref{tab:exp-eng} reports results.
In this experiment, Data Augmentation approach consistently and significantly improves upon
Baseline strategy by 3.8pp, 3.1pp, and 3.0pp with Linear, BiLSTM, and ELMo models 
respectively on Original test set of Twitter, demonstrating much strengthened
generalization power
when the test data is noisier than the training data.

In order to understand the results, we examined some samples from the dataset.
Indeed, on a sentence like `OHIO IS STUPID I HATE IT',
BiLSTM model trained with Baseline strategy was unable to identify `OHIO' as a location although the
state is mentioned fifteen times in the training dataset of CoNLL 2003-English as `Ohio'.
BiLSTM models trained with all other strategies correctly identified the state.
On the other hand, on another sample sentence
`Someone come with me to Raging Waters on Monday',
BiLSTM models from Baseline and Data Augmentation
strategies were able to correctly identify `Raging Waters' as a location
thanks to the proper capitalization,
while the model from Caseless strategy failed on the entity due to its ignorance of orthographic information.


\section{Conclusion}

We proposed a data augmentation strategy for improving robustness of NER models to capitalization errors. 
Compared to previous methods,
data augmentation provides competitive robustness 
while not sacrificing its performance on well-formed text,
and improving generalization to noisy text.
This is consistently observed across models, languages, and dataset sizes.
Also, data augmentation does not require additional language-specific resource, and is trivial to implement
for many natural languages. Therefore, we recommend to use data augmentation
by default for training NER models, especially when 
characteristics of test data are little known a priori.



\bibliography{acl2019}
\bibliographystyle{acl_natbib}

\appendix

\end{document}